# Transformer-Based Low-Resource Language Translation: A Study on Standard Bengali to Sylheti

Mangsura Kabir Oni
Faculty of Computer Science and Engineering,
University of Brahmanbaria, Bangladesh,
mangsurakabir@uob.edu.bd

Tabia Tanzin Prama
University of Vermont, Burlington,
VT 05405, USA
tprama@uvm.edu

*Abstract*— **Machine Translation (MT) has advanced from rule-based and statistical methods to neural approaches based on the Transformer architecture. While these methods have achieved impressive results for high-resource languages, low-resource varieties such as Sylheti remain underexplored. In this work, we investigate Bengali-to-Sylheti translation by fine-tuning multilingual Transformer models and comparing them with zero-shot large language models (LLMs). Experimental results demonstrate that fine-tuned models significantly outperform LLMs, with mBART-50 achieving the highest translation adequacy and MarianMT showing the strongest character-level fidelity. These findings highlight the importance of task-specific adaptation for underrepresented languages and contribute to ongoing efforts toward inclusive language technologies.**

## I. INTRODUCTION

Machine Translation (MT) is the automatic process of converting text from a source language into a target language without human intervention. Over the years, MT has evolved from early rule-based systems to data-driven approaches and, most recently, neural architectures. Rule-Based Machine Translation (RBMT) depended on manually designed linguistic rules and bilingual dictionaries, but such systems were difficult to scale and required extensive expert effort [1]. Statistical Machine Translation (SMT) later reduced manual intervention by learning from bilingual corpora, achieving more fluent translations but struggling with long-range dependencies and semantic adequacy. The shift to Neural Machine Translation (NMT) represented a major advance, leveraging sequence-to-sequence learning with attention to capture contextual relationships more effectively [2].

The introduction of the Transformer architecture further transformed MT by employing self-attention to process long sequences efficiently, outperforming recurrent and convolutional models in both speed and accuracy [3]. Transformer-based systems have since become the backbone of modern MT, enabling the development of multilingual models such as mBART-50, MarianMT, and No Language Left Behind (NLLB-200), which support translation across many languages, including low-resource pairs [8], [9]. In parallel, large language models (LLMs) have expanded the scope of MT by combining massive pretraining with few-shot learning [12]. Models such as GPT-4o, DeepThink_R1, Gemini 2.5 Flash, Claude Sonnet 4, and Sonar demonstrate translation capabilities, though their performance varies widely across languages. Despite these advances, low-resource languages continue to pose significant challenges. Bengali (Bangla) is spoken by over 270 million people worldwide and benefits from moderate computational resources, including corpora and pretrained models [4], [5], [6]. In contrast, Sylheti—spoken by approximately 11–15 million people in Bangladesh, India, and diaspora communities— remains critically underrepresented in NLP research [7]. Although often regarded as a dialect of Bengali, Sylheti differs substantially in phonology, lexicon, and grammar, reducing mutual intelligibility. The lack of standardized orthography, with variations between Bengali script and Romanized writing, further complicates computational processing.

This study focuses on Bengali–Sylheti MT as a representative low-resource task. The objectives are threefold: (i) to create a high-quality parallel corpus for this language pair, (ii) to evaluate fine-tuned transformer-based models (mBART-50, MarianMT, NLLB-200) against zero-shot LLMs, and (iii) to analyze strengths and limitations of each approach. By addressing these objectives, the work aims to contribute to digital inclusivity and support the preservation of Sylheti in the era of multilingual AI technologies.

## II. LITERATURE REVIEW

Machine Translation (MT) has evolved through successive paradigms, each addressing limitations of its predecessors while introducing new capabilities. Early rule-based and example-based systems laid the foundation by leveraging linguistic expertise and explicit grammar rules, yet their applicability was constrained by labor-intensive knowledge engineering and limited scalability [11]. The advent of statistical MT (SMT) marked a shift toward data-driven modeling, where large parallel corpora and probabilistic frameworks enabled more robust translations. Notable work by Brants et al. [12] demonstrated the efficacy of large-scale language models and advanced smoothing strategies, highlighting the growing importance of computational efficiency and corpus size. Neural approaches have since dominated MT research, offering superior fluency and generalization. Techniques such as subword segmentation, back-translation, and Transformer architectures have substantially improved performance in low-resource settings. For instance, Goyal and Sharma [13] showed that applying these strategies in Hindi–English NMT enhanced handling of morphological and syntactic variability. Complementing these advancements, Gatt and Krahmer [14] emphasized that natural language generation research offers valuable insights for MT, particularly in sentence planning and evaluation. Comparative studies, such as Bentivogli et al. [15], reveal that while neural MT reduces morphology and reordering errors relative to phrase-based SMT, challenges remain in translating long and complex sentences. Recent innovations have further expanded the scope of MT. Multimodal approaches integrating text, image, and speech have been surveyed by Sulubacak et al. [16], illustrating the potential of cross-modal contextual information. Incremental subword strategies [17] and factored NMT for morphologically rich languages [18] underscore the importance of fine-grained tokenization and linguistic features in improving translation quality, particularly for low-resource languages. Focusing on Bangla MT, a progression mirroring global trends is evident. GRU-based encoder–decoder systems [19] initially outperformed LSTM architectures, while subsequent Transformer-based models with back-translation [20,22] significantly surpassed both RNN and SMT baselines. Subword modeling further enhanced English–Bangla translation [21], demonstrating the persistent value of token-level innovations. Complementary example-based approaches leveraging lexical resources like WordNet [23] highlight avenues for semantically informed, computationally efficient MT, indicating that hybrid strategies can provide practical benefits alongside neural methods. Despite these advances, the Bengali–Sylheti language pair remains critically underexplored. Scarce standardized corpora, orthographic variation, and limited linguistic resources pose substantial barriers. Despite progress in Bangla MT, the Bengali–Sylheti pair remains underexplored due to scarce corpora and orthographic variation. Prama and Anwar [24] addressed this gap with a Sylheti-to-Bangla NMT system, where BiLSTM achieved the best BLEU-1 score (57.4%), though challenges of limited data and resource constraints persist.

To address this gap, the present study evaluates pretrained multilingual models (mBART-50, MarianMT, NLLB) and state-of-the-art large language models (GPT-4o, DeepThink_R1, Gemini 2.5 Flash, Claude Sonnet 4, Sonar) on a curated Bengali–Sylheti dataset. By synthesizing insights from prior work, this effort aims to extend MT research to low-resource Indic languages, supporting more inclusive and robust NLP systems.

## III. BACKGROUND STUDY

### A. Neural Machine Translation (NMT)

Neural Machine Translation (NMT) is an end-to-end framework mapping source to target sentences using deep neural networks, integrating encoding, decoding, and alignment. Early NMT relied on RNNs with attention to focus on relevant input parts [9], but struggled with long-range dependencies. The Transformer [3] addressed these issues via self-attention, enabling parallel training and superior performance, forming the foundation for modern large language models.

### B. Transformer

The Transformer overcomes sequential limits of RNNs by using self-attention to model dependencies in parallel [3]. Its encoder–decoder design maps input sequences into contextual representations, with the decoder generating outputs auto-regressively. Key components include Scaled Dot-Product Attention, Multi-Head Attention, feed-forward layers, residual connections, and positional encodings. This fully parallelizable architecture (shown in figure 1) captures global context effectively and underlies modern LLMs and MT systems.

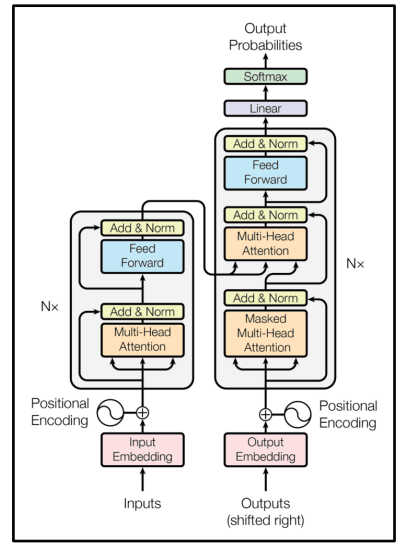

**Figure 1:** Encoder–Decoder Design of the Transformer Model [3]

### C. Encoder-Decoder Models

Encoder–decoder models process input sequences into continuous representations and generate outputs token by token [8]. Attention mechanisms [9] allow the decoder to focus dynamically on relevant input parts, improving translation quality. Transformers [3] replaced recurrence with self-attention while maintaining the encoder–decoder structure, enabling fully parallel computation and forming the basis for current neural architectures.

**mBART-50:** mBART-50 extends mBART-25 to 50 languages, using a 12-layer Transformer encoder and decoder with 16 attention heads and 1024 hidden dimensions [3]. LayerNorm embeddings improve stability, while SentencePiece tokenization ensures consistent multilingual input processing.

**MarianMT:** MarianMT is a Transformer-based multilingual NMT system [3], typically with 6 encoder and 6 decoder layers, each having 8 attention heads and 512 model dimensions. It employs SentencePiece tokenization and is optimized for translation speed and low-resource scenarios.

**NLLB-200:** NLLB-200 supports 200+ languages using Transformer encoder–decoder architecture [3]. It integrates Sparsely Gated MoE layers for efficient scaling and low-resource performance. Input sequences are tokenized via

SentencePiece, embedded, and processed through attention and feed-forward layers to generate translations.

### D. Large Language Models

LLMs capture long-range dependencies using deep self-attention architectures [12]. They factorize sequence probabilities autoregressively, generalizing n-gram models, and rely on distributed computation and smoothing strategies for efficiency. LLMs extend the encoder–decoder paradigm for high-quality MT, summarization, and question answering.

**GPT-4o:** GPT-4o (OpenAI) handles multimodal inputs, using sparse attention and Transformer layers for reasoning, dialogue, and content generation [10].

**DeepThink_R1:** DeepThink_R1 emphasizes long-context reasoning with hierarchical attention on a Transformer backbone, suited for knowledge extraction and analytical tasks [25].

**Gemini 2.5 Flash:** Gemini 2.5 Flash (Google DeepMind) is a multimodal model using hybrid dense-sparse attention for long sequences, optimized for dialogue, reasoning, and multi-turn conversations [26].

**Claude Sonnet 4:** Claude Sonnet 4 (Anthropic) is a conversational LLM with enhanced attention and context management, supporting coherent multi-step instructions and reasoning [27].

**Sonar:** Sonar is a Transformer-based LLM optimized for code comprehension, program synthesis, and automated explanations, leveraging augmented attention for long-range dependencies [28].

### E. Evaluation Metrics for Machine Translation

**BLEU:** BLEU evaluates n-gram overlap between candidate and reference translations, penalizing brevity [2,3,12]. It is widely used for neural and statistical MT due to its simplicity and correlation with human judgment [3,15].

**chrF:** chrF computes character n-gram F-scores, capturing fine-grained morphological differences [14,15]. It is particularly effective for morphologically rich, low-resource languages like Bengali and Sylheti [15].

## IV. DATASET DESCRIPTION

The dataset for this study was meticulously compiled to support neural machine translation (NMT) between Sylheti and Standard Bengali. Sylheti, spoken by over 11 million people across Bangladesh, India, and diaspora communities, has historically been underrepresented in digital and written resources, creating significant challenges for computational modeling. To address this, a bilingual parallel corpus (shown in Table 1) of 5,002 sentence pairs was constructed, covering both spoken and written varieties. Data were sourced from diverse domains, including newspapers (e.g., Prothom Alo, The Daily Star, Ittefaq), social media platforms (Facebook, Twitter, Instagram, YouTube), contributions from native speakers, literature and folklore, drama and entertainment dialogues, and published books. The corpus was analyzed at multiple linguistic levels, containing 21,132 unique words (9,729 Sylheti; 11,403 Standard Bengali), 10,340 clauses, and 10,004 aligned sentences. Sample word, clause, and sentence alignments illustrate the lexical, syntactic, and semantic correspondences across the two languages. This dataset is being prepared for formal publication to provide a reusable and reliable resource for low-resource language processing, multilingual NLP, and inclusive AI research.

**Table 1:** Sample Word, Clause, and Sentence Alignments of the Sylheti–Standard Bengali Parallel Corpus

| | Category | |
|---|---|---|
| | Sylheti Language | Standard Bangla Language |
| **Word** | আথথা | হঠাং |
| **Clause** | আইলর ঘাস | আঁইলের ঘাস |
| **Sentence** | শীতর হাওয়া আমারে ছুইয়া যায় | শীতের হাওয়া আমাকে ছুঁয়ে যায় |

## V. METHODOLOGY

This section presents the methodology for building and evaluating Bengali–Sylheti machine translation systems. The workflow has two main branches: (i) zero-shot translation using large language models (LLMs) and (ii) fine-tuning transformer-based NMT models. Both branches use a cleaned parallel corpus and follow systematic steps of data preparation, model training or prompting, and evaluation. The aim is to compare the effectiveness of general-purpose LLMs with specialized NMT models for low-resource translation. Figure 2 illustrates the complete workflow from corpus preparation to evaluation and comparative analysis.

### 1. Cleaned Corpus

The Bengali–Sylheti parallel corpus serves as the foundation for both branches. The raw dataset underwent rigorous cleaning to remove duplicates, inconsistent formatting, and noisy entries. Sentences were normalized to ensure uniform spelling, punctuation, and tokenization, which is particularly important for Sylheti due to its non-standardized orthography. The cleaned corpus was split into training (80%), validation, and testing (20%) subsets to guarantee independent evaluation. This structured dataset provided a reliable backbone for all subsequent experiments and ensured fair comparisons across models.

### 2. LLM Branch:

The first experimental branch investigated the zero-shot translation potential of LLMs, including GPT-4o, Claude Sonnet, Gemini 2.5, and Sonar. In this branch, Standard Bengali sentences were provided as input prompts, and Sylheti translations were generated based on the models' multilingual pretraining and emergent translation capabilities. Prompt engineering was used to clearly define the source and target languages, improving alignment with the expected Sylheti output.

The primary advantage of this approach is its minimal resource requirement: no fine-tuning or GPU-intensive training is necessary. However, LLMs are not specialized for Sylheti, making this branch a baseline for evaluating general-purpose translation capabilities. Generated outputs were systematically stored and later evaluated using BLEU and ChrF metrics to facilitate a fair comparative analysis against fine-tuned transformer models.

### 3. Preprocessing Pipeline
To ensure high-quality, model-compatible inputs, the Bengali–Sylheti corpus underwent a comprehensive preprocessing pipeline.

**Data Splitting:** The cleaned corpus was divided into 80% training and 20% testing using a fixed random seed to guarantee reproducibility. This consistent split ensured that all models were trained and evaluated on the same data partitions.

**Format Normalization**: Each sentence pair was organized with Bengali as the source language (tagged as bn) and Sylheti as the target language (tagged as syl). The dataset was stored in JSON and TSV formats to maintain compatibility with multiple transformer frameworks and Hugging Face tokenizer utilities. This standardization eliminated ambiguity during training and evaluation.

**Data Augmentation**: To improve model generalization and reduce sparsity, the corpus was enriched using three augmentation strategies:

1. **Back-Translation:** Sentences were translated through an intermediate language and then back to Bengali or Sylheti to create paraphrased variants while preserving semantic meaning.
2. **Synonym Replacement:** Selected words were replaced with synonyms to increase lexical diversity without changing sentence semantics.
3. **Character-Level Perturbation:** Minor character edits simulated orthographic variations common in Sylheti, enhancing model robustness to noisy input.

**Tokenization and Language Tagging**: Pretrained tokenizers converted text into subword units, and language codes were applied to guide model outputs in the intended target language. This step was critical for both multilingual and bilingual models.

**Length Normalization**: Sequences were either truncated to a maximum token length or padded with special tokens to maintain uniform batch sizes. Dynamic batch-level padding was used where applicable to reduce computational waste.

### 1. Transformer Branch: Fine-Tuning Models
The second experimental branch involved fine-tuning three transformer-based NMT models: mBART-50, MarianMT, and NLLB-200. All models were trained on the preprocessed corpus, with identical 80/20 splits to ensure comparability.

**Fine-Tuning mBART-50**: mBART-50 leveraged its multilingual pretraining to adapt to Bengali–Sylheti translation. The model used augmented training data, tokenization, and language tagging (source: bn_IN, target: en_XX). Fine-tuning applied sequence-to-sequence training with label smoothing, dropout, gradient accumulation, and partial layer freezing to optimize performance. Early stopping based on BLEU scores prevented overfitting, ensuring reproducibility and fair comparison with other models.

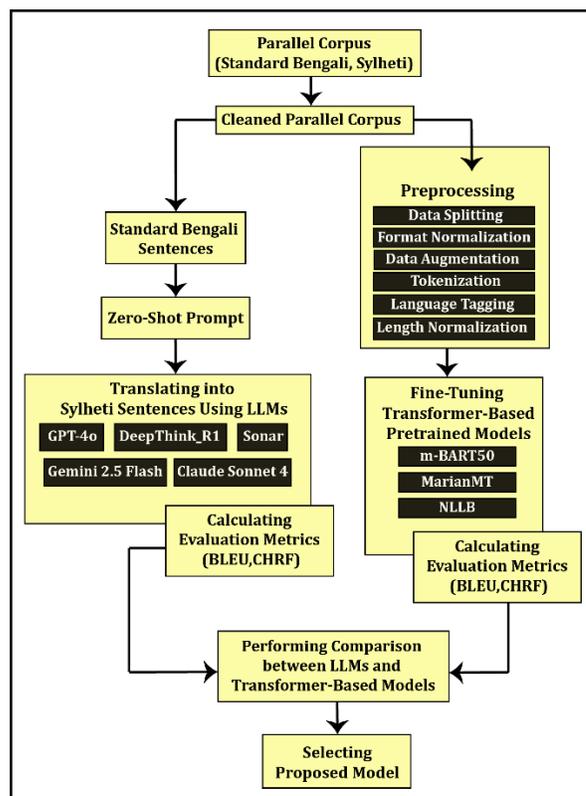

**Figure 2:** End-to-End Methodological Workflow for Bengali–Sylheti MT

**Fine-Tuning MarianMT**: MarianMT, a lightweight, language-pair-focused model, was fine-tuned for 30 epochs with AdamW optimization, FP16 mixed precision, and weight decay regularization. The dataset followed a structured translation schema, ensuring tokenization compatibility. BLEU, ChrF, and METEOR metrics were used for validation, allowing a comprehensive assessment of translation quality.

**Fine-Tuning NLLB-200**: NLLB-200 utilized augmentation techniques, including back-translation, character perturbation, and synonym replacement, to expand the dataset. Tokenization and multilingual language codes (source: ben_Beng, target: sil_Beng) ensured proper handling. Training applied AdamW optimization, weight decay, FP16 precision, and early stopping. BLEU and ChrF metrics guided model selection and evaluation.

All three models shared common strategies—sequence-to-sequence fine-tuning, early stopping, and standard data splits—

while differing in architecture, scale, and augmentation focus. Large multilingual models (mBART-50 and NLLB-200) emphasized embeddings and language codes, whereas MarianMT prioritized resource-efficient adaptation.

### 2. Evaluation Setup:

Both LLM and transformer branches were evaluated using **BLEU** (for n-gram overlap) and **ChrF** (for character-level similarity), particularly suitable for morphologically rich Sylheti. All models were assessed on the same held-out test set to ensure fairness.

### 3. Comparative Framework:

The methodology allowed a structured comparison between general-purpose zero-shot LLMs and task-specific fine-tuned transformers. LLMs offered flexibility and minimal setup, whereas fine-tuned transformers provided higher translation accuracy and consistency. Using identical test data and evaluation metrics, this framework highlighted the trade-offs between model generality, resource requirements, and domain-specific performance.

## VI. EXPERIMENTAL RESULTS

This section presents the evaluation of Bengali–Sylheti machine translation models, focusing on fine-tuned transformer architectures (mBART-50, MarianMT, and NLLB-200) and comparing them with zero-shot performance of five large language models (LLMs). Training was conducted for up to 30 epochs, with performance assessed using BLEU (n-gram precision and adequacy) and chrF (character-level fidelity). Beyond numerical metrics, training dynamics, convergence stability, and qualitative translation behavior were also analyzed to provide a comprehensive understanding of system performance.

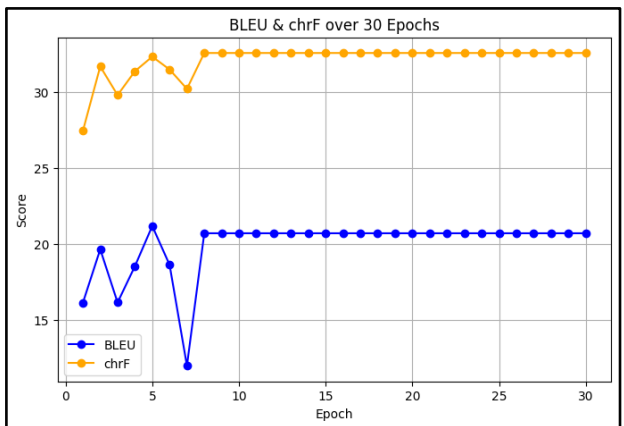

**Figure 3:** mBART-50 Model Performance Metrics Across Training Epochs

### Performance of Fine-Tuned Transformer Models

The three fine-tuned transformer models—mBART-50, MarianMT, and NLLB-200—demonstrated significant improvements over zero-shot LLMs in translating Bengali to Sylheti. Table 2 summarizes their overall performance, highlighting differences in BLEU and chrF scores, training stability, and morphological fidelity. Each model exhibited distinct strengths and learning behaviors, which are discussed below.

**mBART-50:** The mBART-50 model demonstrated rapid initial learning but lacked stability across epochs. As illustrated in Fig. 3, BLEU peaked at 21.17 (epoch 5), but dropped sharply to 11.99 (epoch 7) before recovering to 20.66 (epoch 8). This volatility reflects sensitivity to overfitting on the limited dataset. In contrast, chrF exhibited steadier growth, reaching 32.55 at epoch 8. Validation loss consistently decreased from 9.77 to 9.46 in the early stages, confirming effective optimization before instability set in. These results highlight mBART-50's potential for strong early performance, but they also emphasize the importance of early stopping when applied to low-resource translation tasks.

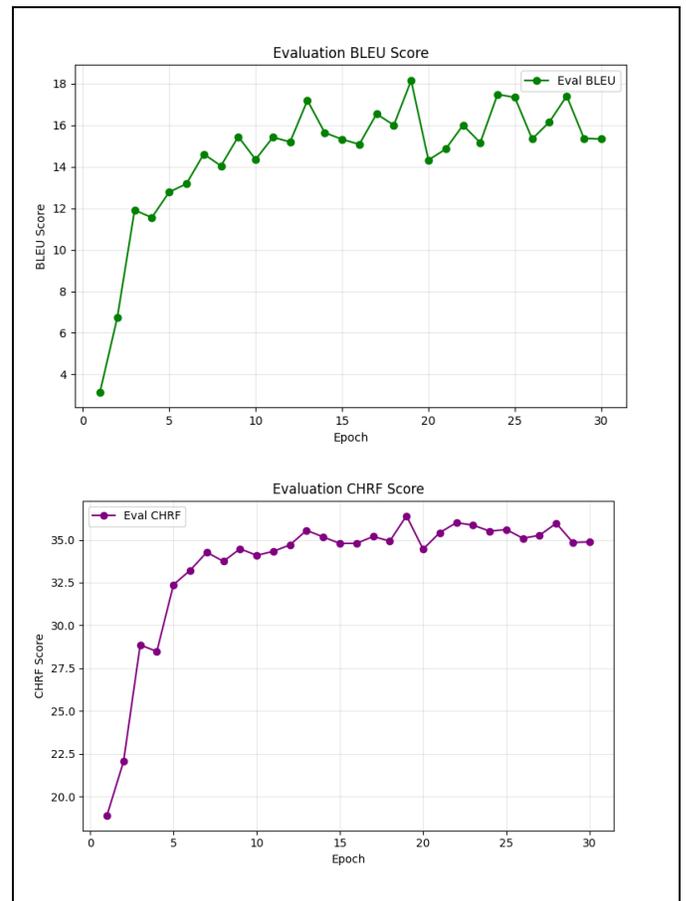

**Figure 4:** MarianMT Model Performance Metrics Across Training Epochs

**MarianMT:** MarianMT achieved the most stable and consistent performance across the 30 training epochs. As shown in Fig. 4, BLEU steadily improved, reaching 18.16 (epoch 19), while chrF attained the highest score in this study at 36.41, indicating superior character-level accuracy. Training loss decreased smoothly from 2.72 to 0.41, and validation loss stabilized around 1.55–1.73 after epoch 10, reflecting effective generalization. MarianMT's steady trajectory and high chrF performance make it particularly effective in capturing Sylheti's morphological and orthographic richness without severe overfitting.

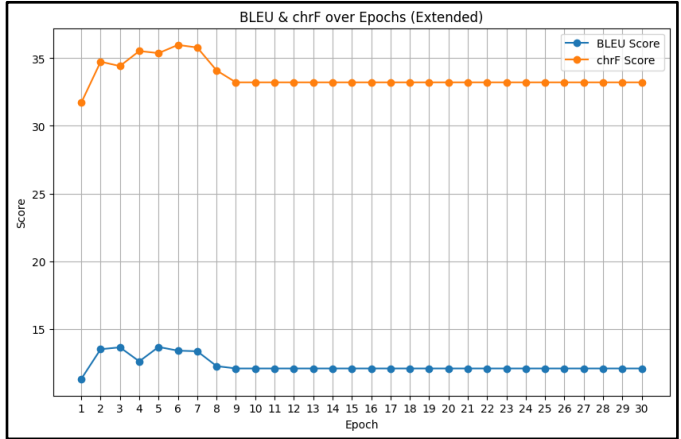

**Figure 5:** NLLB Model Performance Metrics Across Training Epochs

**NLLB-200:** The NLLB-200 model presented the most unstable learning profile. BLEU peaked modestly at 13.65 (epoch 5), while chrF reached 35.98 (epoch 6) before declining. As illustrated in Fig. 5, fluctuations in both metrics prevented stable convergence. Validation loss also trended upward after epoch 5, rising from 0.15 to 0.16, suggesting challenges in adapting the large multilingual architecture to this under-resourced pair. Despite its strong cross-lingual foundations, NLLB-200 appears to require more domain-specific fine-tuning or larger corpora to reach competitive adequacy in Sylheti translation.

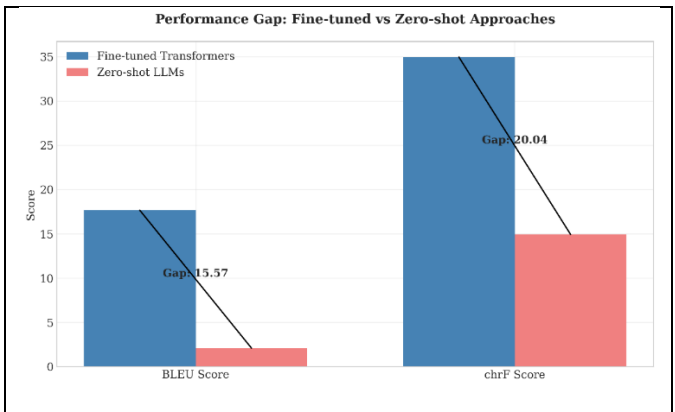

**Figure 6:** Performance Comparison: Fine-Tuned vs. Zero-Shot Translation Approaches

**Zero-Shot Large Language Models:** Zero-shot LLMs performed significantly below the fine-tuned systems. Among the five evaluated models, Gemini 2.5 Flash performed best with BLEU 2.89 and chrF 17.33. GPT-4o and Claude Sonnet 4 scored similarly (BLEU: 1.71–2.10, chrF: 13.97–14.56), while Sonar recorded the lowest performance (BLEU: 1.63, chrF: 13.67). These results confirm that current LLMs lack sufficient exposure to Sylheti during pretraining. Outputs often contained literal Bengali transfers, code-mixing, and orthographic inconsistencies, highlighting the limitations of zero-shot LLM translation for underrepresented languages.

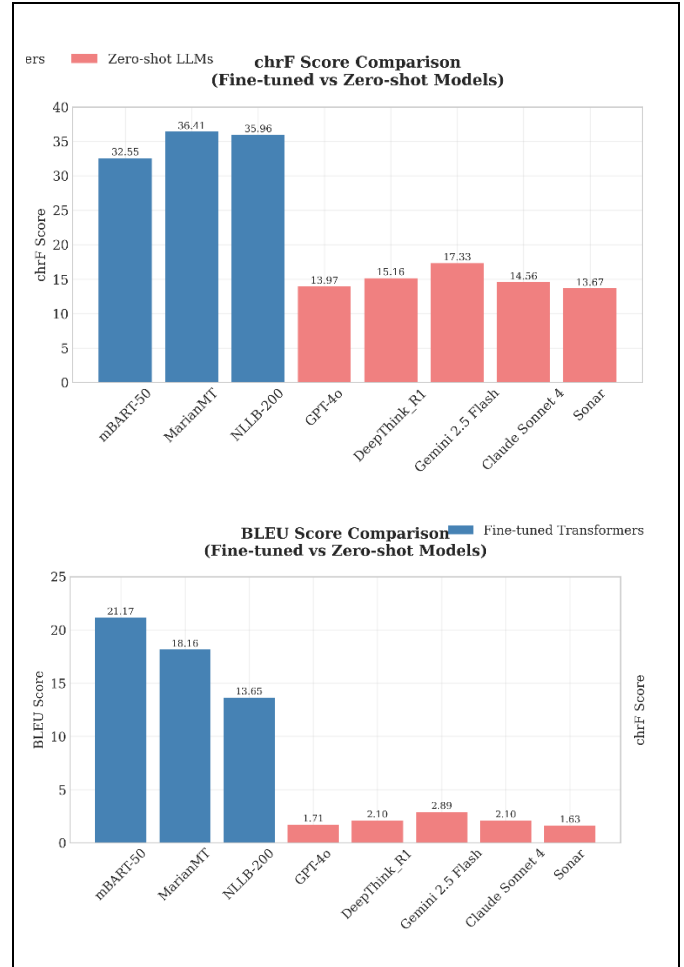

**Figure 7:** Comparative Overview of Best-Achieved BLEU and chrF Scores

**Comparative Analysis:** The comparative evaluation underscores a clear performance gap between the two approaches. As summarized in Fig. 6, fine-tuned transformers outperformed zero-shot LLMs by an average of 15.57 BLEU (17.66 vs. 2.09) and 20.04 chrF (41.31 vs. 21.27).

Among transformers, Fig. 7 shows that: mBART-50 achieved the highest single BLEU score (21.17), excelling in semantic adequacy and word order. MarianMT maintained the most stable trajectory, achieving the best chrF (36.41) and producing morphologically precise outputs. NLLB-200 delivered competitive chrF but consistently underperformed on BLEU, indicating weaker sentence-level adequacy. Correlation analysis (Fig. 8) revealed that BLEU and chrF generally align, though not perfectly. MarianMT, for example, outperformed in chrF relative to BLEU, suggesting it produces character-accurate but slightly less reference-aligned translations.

**Error Analysis and Observations**

Manual inspection revealed that fine-tuned models handled common Sylheti constructs well but struggled with rare words, idiomatic expressions, and proper nouns. MarianMT excelled in verb conjugations and possessive forms, while mBART-50 produced more reference-faithful structures at peak epochs. NLLB-200 frequently exhibited hallucinations and instability. LLM outputs, however, were dominated by literal transliterations from Bengali and inconsistent grammar. Orthographic variation in training data also influenced all models, sometimes yielding alternate but valid spellings penalized by BLEU.

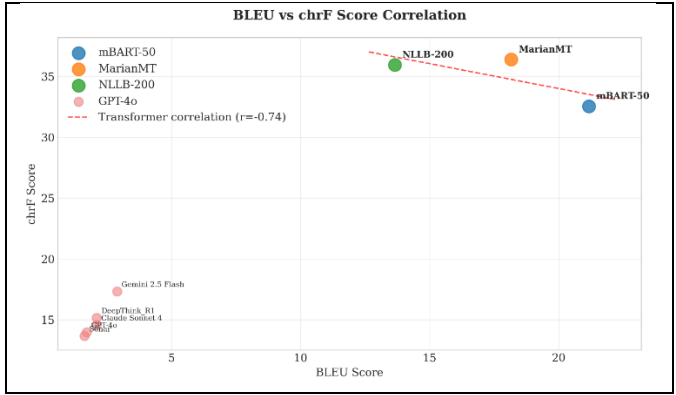

**Figure 8:** Correlation Analysis of BLEU and chrF Scores for All Systems

**Conclusion and Model Recommendation**

The experimental evidence supports the superiority of fine-tuned transformer models over zero-shot LLMs for Bengali–Sylheti translation. Among them:

mBART-50 delivered the highest BLEU (21.17), making it the most semantically accurate and reference-faithful system when training is stopped early. MarianMT achieved the highest chrF (36.41) with steady learning and strong morphological fidelity, making it well-suited for practical deployments requiring robustness. NLLB-200 showed potential but underperformed due to difficulties adapting to a niche low-resource setting. Given BLEU's stronger alignment with overall translation adequacy, mBART-50 is recommended as the primary model. For deployment, we propose implementing early stopping around epoch 5, ensuring peak performance is captured before overfitting occurs. The sharp contrast with LLM performance further emphasizes the necessity of targeted fine-tuning for severely underrepresented languages like Sylheti.

## I. CONCLUSION

This study evaluated Bengali–Sylheti machine translation using three fine-tuned transformer models (mBART-50, MarianMT, and NLLB-200) alongside five zero-shot large language models. The findings indicate that fine-tuned models consistently outperform zero-shot LLMs, achieving over 15 BLEU and 20 chrF improvements on average. Among transformer systems, mBART-50 reached the highest BLEU score (21.17), demonstrating strong semantic adequacy though with training instability requiring early stopping. MarianMT achieved the best chrF (36.41) with stable convergence, effectively capturing Sylheti's morphological and orthographic patterns, while NLLB-200 showed limited adaptability in this low-resource setting. In contrast, zero-shot LLMs performed poorly (BLEU ≤ 2.89, chrF ≤ 17.33), frequently producing literal or code-mixed translations. Overall, mBART-50 with early stopping emerges as the most effective system, while MarianMT provides a stable alternative. Future work will explore data augmentation, orthographic normalization, and hybrid approaches to further improve translation quality.

**Table 2:** Performance Summary of Bengali–Sylheti Translation Models

| Model | BLEU (Best) | chrF (Best) | Peak Epoch | Training Loss Trend |
|---|---|---|---|---|
| mBART-50 | 21.17 | 32.55 | 5 | Decreased early, then unstable |
| MarianMT | 18.16 | 36.41 | 19 | Smooth and stable decrease |
| NLLB-200 | 13.65 | 35.98 | 5–6 | Fluctuating, weak adaptation |
| Gemini 2.5 Flash | 2.89 | 17.33 | — | Zero-shot, unstable outputs |
| GPT-4o | 2.10 | 14.56 | — | Zero-shot, literal transfers |
| Claude Sonnet 4 | 1.71 | 13.97 | — | Code-mixed, unstable grammar |
| Sonar | 1.63 | 13.67 | — | Weakest performance |

## II. LIMITATIONS AND FUTURE WORK

Although the findings highlight the effectiveness of fine-tuned transformer models for Bengali–Sylheti translation, several limitations remain. The dataset size (5,002 parallel sentences) restricts the models' capacity to generalize across diverse syntactic structures, stylistic variations, and domain-specific expressions. In addition, orthographic inconsistencies in Sylheti introduce noise, leading to training instability, particularly in models like mBART-50. Another limitation is the reliance on automatic evaluation metrics such as BLEU and chrF, which may not fully capture the linguistic richness or cultural nuance of Sylheti. Future research should therefore focus on expanding the dataset through community-driven contributions and data augmentation strategies. Incorporating orthographic normalization could improve consistency and reduce variability during training. Hybrid approaches that combine the strengths of pre-trained LLMs with fine-tuned NMT models may also enhance translation robustness in low-resource settings. Finally, incorporating human evaluation will provide a more comprehensive assessment of translation adequacy, fluency, and cultural alignment.

### ACKNOWLEDGMENT

The authors would like to thank the native Sylheti speakers and community contributors who provided invaluable linguistic insights that guided the creation of the Sylheti–Bengali parallel corpus. The dataset was primarily collected and curated by the authors themselves. The authors also acknowledge the role of open-source frameworks and tools, including Hugging Face Transformers, MarianMT, mBART-50, and NLLB-200, which enabled the fine-tuning and evaluation of translation models. Their combined contributions have been critical in advancing research on low-resource language processing.